\documentclass[reprint,
 amsmath,amssymb,
 aps,
 prl,
floatfix,
]{revtex4-1}
\usepackage{xcolor}

\usepackage{graphicx}
\usepackage{dcolumn}
\usepackage{bm}
\usepackage{hyperref}
\usepackage{braket}

\begin{document}

\newcommand{\comment}[1]{}

\preprint{APS/123-QED}


\title{Stochastic resonance neurons in artificial neural networks}

\author{E. Manuylovich, D. Arg\"uello Ron, M. Kamalian-Kopae, S. K. Turitsyn }

\address{Aston Institute of Photonic Technologies, Aston University, Birmingham B4 7ET, UK}
\date{\today}


\begin{abstract}
 We propose a new type of artificial neural networks using stochastic resonance as a dynamic nonlinear node. New design
 enables significant reduction of the required number of neurons for a given performance accuracy. We also observe that such  system is more robust against the impact of noise compared to the conventional analogue neural networks.  

\end{abstract}


\maketitle


%
%
 Artificial neural networks (ANNs)  are capable of solving certain problems using only the accessible sets of training examples, without knowledge of the underlying systems responsible for the generation of this data (see e.g.~\cite{ANN4,ANN1,ANN5} and references therein). In scientific and engineering applications ANNs can be employed as a non-linear statistical  tool that  learns low-dimensional representations from complex data and  uses this to model nontrivial relationships between inputs and outputs. 
 
 Ability of ANNs to predict and approximate from given data is linked to the effective dimensionality - the number of independent free parameters in the model.  The approximation capability of the ANNs is quantified by the universal approximation theorems (see for details, e.g.~\cite{UAT01}). Complexity of the ANNs  plays a critical role in the trade-off between their performance and accuracy of the predictions on the one side and power consumption and speed of operation on the other.  Many modern applications of ANNs requires over-parametrized models in order to ensure optimal performance, which results in high computational complexity and corresponding increased power consumption.

Power efficiency can be improved using  physically implemented (not necessarily digital) neural networks, for instance, designed from layers of controllable physical elements \cite{PINN01}. Physics-inspired or physics-informed analogue neural networks, capable to combine data processing with the knowledge of the underlying physical systems embedded into their architecture \cite{PINN01,PINN02,PINN03,PINN04,PINN05} is a fascinating area of  research offering, in particular, a potential pathway to power-efficient ANNs. There is, however, an essential trade-off between improvements in energy efficiency and susceptibility to noise in the analogue networks. The  fundamental challenge in non-digital systems is the accumulation of noise originating from analogue components~\cite{Analog01,semenova2019fundamental, zhou2020noisy,semenova2022understanding}. To overcome this challenge and unlock full potential of analogue ANNs, it is required to develop systems with capability of absorbing, converting and transforming noise. In this Letter we propose a new design of  artificial neural network  with stochastic resonance  \cite{RevModPhys.70.223} used as a network node - ANN-SR.

 The  robustness of the ANNs operation in the presence of noise depends on the properties of the nonlinear activation function~\cite{semenova2022understanding}. Instead of using a conventional static nonlinear element, we employ a dynamical system with bi-stable features (see Fig 1) that can make a positive use of the noise -  the stochastic resonance (SR). Note that there are various experimental implementations of SRs that pave the way to new interesting physically implementable designs of ANNs~\cite{dodda2020stochastic,singh2001stochastic}.  We demonstrate that ANN-SR can significantly reduce the computational complexity and the required number of neurons for a given prediction accuracy. Moreover, ANN-SR performance is more robust against noise in the case of training on noisy data.

\begin{figure}[ht!]
\includegraphics[width=.98\linewidth]{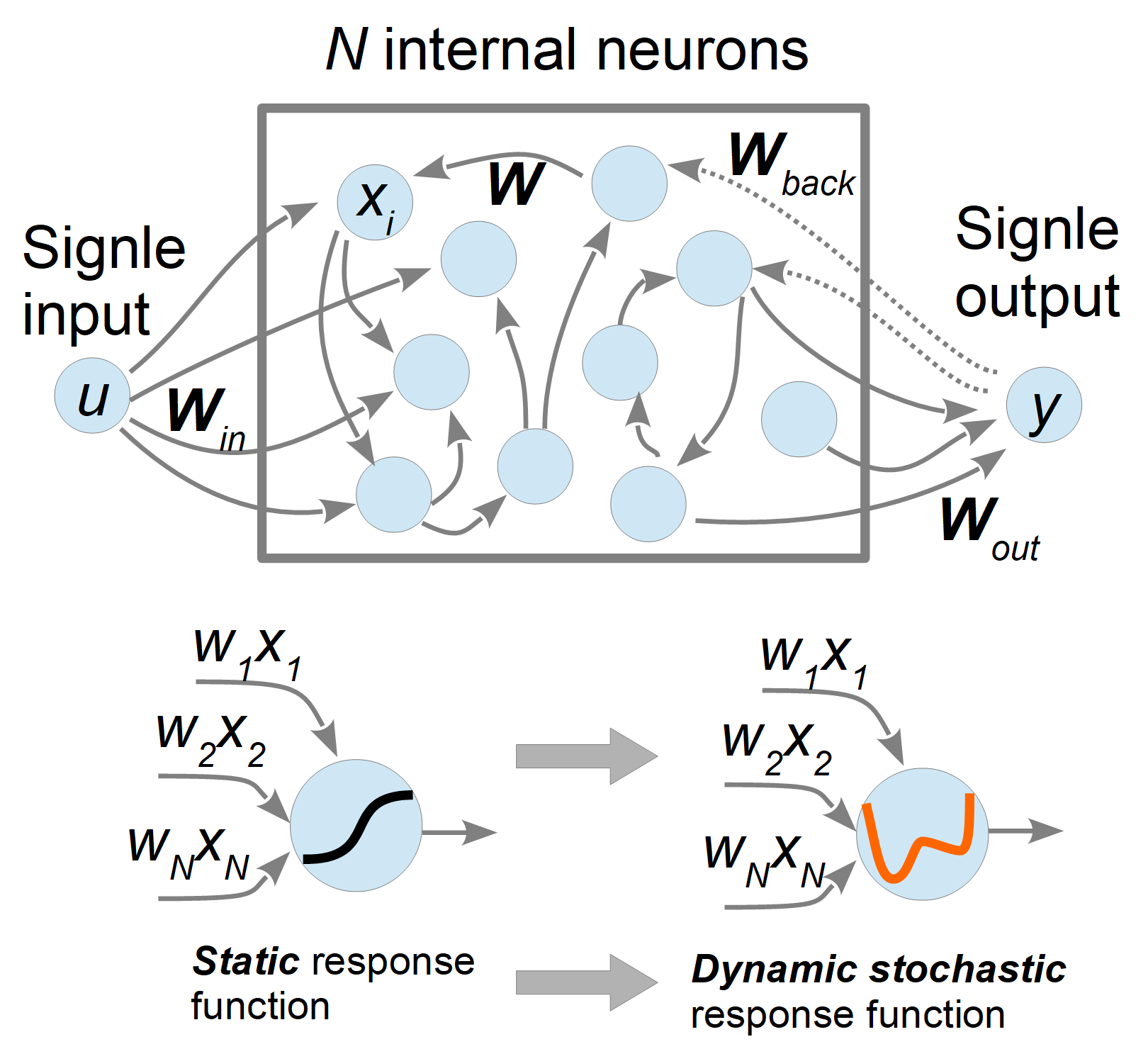}
\caption{Scheme of the considered ESN and the proposed use of dynamic stochastic response functions instead of static ones.}
\label{fig: ESN scheme}
\end{figure}
We consider here, as an illustration of our general idea, a particular recurrent neural network schematically shown in Fig.~\ref{fig: ESN scheme}, known as echo state network (ESN)~\cite{Jaeger2004}, however, the idea of SR neurons can be easily extended to other architectures such as, e.g. deep neural networks. ESNs are fixed recurrent neural networks that constitute a "reservoir" with multiple (fixed) internal interconnections providing a complex nonlinear multidimensional response to an input signal. An output signal is obtained by training linear combinations only of these read-out ESN responses.
The ESN has an internal dynamic memory. The internal connections of the ESN are randomly set up. During the learning procedure the training sequence is fed into the ESN and after an initial transient period the neurons start showing some variations of the fed signal or "echoing" it. After feeding the training sequence to the ESN the readout weights are calculated.

To explain mathematics behind our idea, we consider the classical ESN model \cite{Jaeger2004}:
\begin{equation} 
\begin{aligned}
& x_{n+1} = f(\mathbf{W}_{in}u_{n+1} +  \mathbf{W}x_{n} + \mathbf{W}_{back}y_{n+1}) &\\
\end{aligned}
\label{eq: ESN evolution step}
\end{equation}
where $u \in \mathbb{R}^{M}$ is the current state of the ESN's input neuron, $x_{n} \in \mathbb{R}^{N \times 1}$ is the vector with dimension $N$ corresponding to the network internal state, matrix 
$\mathbf{W}^{in} \in \mathbb{R}^{N \times 1}$ is the map of the input $u$ to the vector of dimension $N$, matrix $\mathbf{W} \in \mathbb{R}^{N \times N}$ is
the map of the previous state of the recurrent layer, and $f(x)$ is the nonlinear activation function. The recurrent layer consists of $N$ neurons, and the input information $u$ contains a single element at each instance of time.
The linear readout layer $\mathbf{W}_{out}$  is defined as
\begin{equation} \label{eq: ESN readout layer equation}
y_{n+1} = \mathbf{W}_{out} x_{n+1}  
\end{equation}
where vector $W_{out} \in \mathbb{R}^{1 \times N}$ maps the internal state of the ESN to a single output. 

In the classical approach the nonlinear activation function $f(x)$ is a sigmoid or hyperbolic tangent function~\cite{Jaeger2004}. We replace in the model \ref{eq: ESN evolution step} the analytical nonlinear function $f(x)$ with a stochastic ordinary  differential equation known as stochastic resonance~\cite{nb2021noise,anishchenko1999stochastic}. The SR phenomenon has been studied in a wide range of physical systems, such as climate modelling, electronic circuits, neural models, chemical reactions, and photonic systems (see e.g.~\cite{RevModPhys.70.223,harmer2002review,nb2021noise,anishchenko1999stochastic,RevModPhys.85.421}). This type of dynamics can be represented using a bi-stable system with two inputs: a coherent signal and a noise~\cite{balakrishnan2019chaosnet, harmer2002review}. 
A standard example of the SR model reads:
\begin{equation}
 \label{eq: SR analytical equation}
\frac{d \xi(t)}{d t} = -\frac{dU_{0}(\xi)}{d\xi} + s(t) + D\sigma (t).
\end{equation}
Here $s(t)$ represents the input signal that will be transformed by the dynamical system into the output signal $\xi(t)$, $\sigma(t)$ is the Gaussian noise with zero mean and variance of 1, where $D$ is the noise amplitude. The stationary potential function $U_{0}$ together with the time-dependent input signal $s(t)$ define the time-dependent landscape of internal evolution of the dynamical system and form the time-dependent tilted potential $U(\xi,t):$ 
\begin{equation}
 \label{eq: SR time-dependent potential}
U(\xi,t) = U_{0}(\xi) + \xi s(t)
 \end{equation}
 Let us  consider a symmetric bi-stable stationary potential well $U_{0}$:
\begin{equation}
 \label{eq: SR potential}
U_{0}(x) = -\alpha\cdot \frac{x^{2}}{2} + \beta\cdot \frac{x^{4}}{4}
 \end{equation}
This model is a limit of  the heavily dumped  harmonic oscillator, which represents the movement of a particle in the time-dependent bi-stable potential $U(x,t)=U_{0}(x) - x s(t)$ \cite{harmer2002review,Mingesz}. When $\alpha>0$, the potential is bi-stable with two stationary points $x_{s} = \pm \sqrt{\alpha/\beta}$ and a barrier of value $\Delta U = \alpha^2/(4\beta)$. 

\comment{Figure~\ref{fig: tilted potential} shows various potential functions $U(x)$ for different input signals $s$. These landscapes define the evolution of the internal state of an SR system.
\begin{figure}[ht!]
    \includegraphics[width=8cm]{fig02. SR as a nonlinear function.png}
    \caption{Landscape of internal evolution of the SR function.}
    \label{fig: tilted potential}
\end{figure}
}

Noise is a requisite ingredient in defining SR operation.  The utilization of noise as a tool for improving the performance of deep learning algorithms has been widely investigated~\cite{nagabushan2016effect, bishop1995neural}.
Noise has been used to combat adversarial attacks~\cite{colorednoise, parametricnoise}, as a regularization technique~\cite{gaussiannoise} and as a way to increase the noise robustness of ANNs that are implemented in analogue hardware~\cite{zkmw20, zhou2020noisy}. Furthermore, different noise injection techniques and different scenarios with adding noise to the input, weights or activation function are explored in recent studies~\cite{parametricnoise,zkmw20}. Noise increases the dimensions of the latent feature space \cite{Noise01,Noise02} and the probability of a more chaotic operation. 
\comment{
 \begin{equation} \label{eq:ANN1}
y' = f(\mathbf{w}\cdot \mathbf{x}  + \mathbf{b}) + \mathbf{\eta}
\end{equation}
$S_{k} -> x_{n}$
\begin{equation} \label{eq:ANN1}
s_{k} = \mathbf{w_{n}}\cdot \mathbf{x_{n}}  + \mathbf{b_{k}}
\end{equation}
\begin{equation} \label{eq:ANN1}
 y' = f((\mathbf{w}+\mathbf{\eta})\cdot \mathbf{x}  + \mathbf{b})
 \end{equation}
\begin{equation} \label{eq:ANN1}
y' = f(\mathbf{w}\cdot \mathbf{x}  + \mathbf{b} + \mathbf{\eta})
 \end{equation}
 }
Here we demonstrate that under certain conditions the SR function is superior compared to the classical sigmoid node as a nonlinear activation function. We compare the prediction accuracy and computational complexity and show that SR outperforms the classical approach in both of these aspects.

The standard neuron collects a linear combination of incoming values, applies a nonlinear function to it and passes the result further. The proposed SR neuron is different: it has its own internal state $\xi(t)$, which constitutes memory. This internal state together with the incoming signals from other neurons are used to form the output. The SR neuron collects a linear combination of incoming weights $s$ which modifies the potential function $U$. Then Eq.~(\ref{eq: SR analytical equation}) is integrated on some time interval $\Delta t$ to obtain the next internal state of the SR neuron. This new internal state $\xi(t+\Delta t)$ is the output value of SR neuron.

To evaluate the performance of the SR as the nonlinear function we used a classical problem of predicting the Mackey-Glass (MG) series $q(t)$ (see Supplementary materials). For training and testing the described approach, in addition to the MG series we used a noisy version of the MG series.\comment{ $q_\eta$:
\begin{equation} 
\begin{aligned}
& q_\eta(t)  = q(t) + \mathcal{N}(0,\eta).\\
\end{aligned}
    \label{eq: noisy MG equation}
\end{equation}
}
A part of the MG series is used to train the ESN and the next several tens of points of the MG series are compared to the output of the freely running ESN to estimate the prediction accuracy. During the training and testing, the nonlinear response of each SR neuron is calculated as a result of the integration of Eq.~(\ref{eq: SR analytical equation}) with weighted sum of input neurons as the variable $s$ in Eq.~(\ref{eq: SR analytical equation}) and previous internal state of the neuron as initial condition. As the training and test sequences are parts of the MG series they are time-dependent, therefore, we integrate Eq.~(\ref{eq: SR analytical equation}) over time with $s$ as a parameter. The integration interval $\Delta t$ is equal to the time step of the MG series and we set the initial value of the SR function as $\xi(t=0)=\mathcal{N}(0,1)$. On each time step we take the current value of the function $\xi_n$ and  calculate the next value by integrating Eq.~(\ref{eq: SR analytical equation}) with $\xi_n$ as an initial value: 
\begin{equation} \label{eq: SR RK2 output}
f(s_n)\equiv\xi_{n+1}\!=\!\xi_{n}\!+\!\left[\alpha(\xi\!-\!\xi^{3})\!+\!D\sigma(t)\right]\!\cdot\Delta t+s_n\cdot\Delta t
\end{equation}

This is done whenever we need to calculate the nonlinear response function $f(s)$ using a 2nd order Runge-Kutta (RK2) also known as the Euler method.

\comment{In this way, t}The output of the nonlinear response function is calculated as a trajectory of a system described by the SR equation with input values as an external force applied to the system.
The internal state of the SR system is described by a vector $\xi$ that evolves with time. For a given initial state $\xi_n$ and input value $s_n$, the output of the SR nonlinear function can be calculated as the next internal state $\xi_{n+1}$ separated from the previous state by time $\Delta t$. 
Note that there are two sources of noise: (i) implementation noise coming from the active nature of the nonlinear node, and (ii) the noisy data, both practically important and considered in this work.

In the current work we use the proposed approach to design an ESN and estimate its accuracy depending on the number of neurons and noise amplitude in the training data. We compare the designed ESN-SR with the classical ESN with sigmoid activation function in terms of accuracy and computational complexity.

We estimated the computational complexity of ESNs as the number of multiplications needed to perform a single evolutionary step described by Eq.~(\ref{eq: ESN evolution step}) and Eq.~(\ref{eq: ESN readout layer equation}).

\comment{

According to these equations the linear part of ESN evolution includes 4 matrix multiplications of size $N\!\!\times\!\!1$, $N\!\!\times\!\! N$, $N\!\!\times\!\!1$, and $1\!\!\times\!\!N$ giving a total of
\begin{equation} 
\begin{aligned}
& Q_{\text{linear}} = N^{2}+3N
\end{aligned}
    \label{eq: CC of linear part}
\end{equation}
multiplications per 1 step of linear part of the ESN evolution. This is also the total number of multiplications for the classical approach for the nonlinear activation function such as sigmoid or tanh is obtained using lookup table with no computational complexity.

On the other hand, for the SR nonlinear function, the values of $\alpha(\xi-\xi^{3})+D\sigma(t)$ or $\left[\alpha(\xi_n-\xi_n^{3})+D\sigma(t)\right]\cdot\Delta t$ can also be calculated using look-up tables and bear no computational burden. However, $s_n \cdot \Delta t$ needs to be calculated and add up $\xi_n$, $\left[\alpha(\xi_n-\xi_n^{3})+D\sigma(t)\right]\cdot\Delta t$, and  $s_n \cdot \Delta t$. So an additional
\begin{equation}     \label{eq: SR RK2 nonlinear CC}
Q_{\text{nonlinear}}^{SR}=N 
\end{equation}
multiplications are needed to calculate the SR nonlinear response function.
}

The total numbers of multiplications for 1 step of calculating evolution of the ESN are:
\begin{equation} 
\begin{aligned}
& Q_{\text{total}}^{\text{classical}}=N^2+3N \\
& Q_{\text{total}}^{SR}=N^2+4N \\
\end{aligned}
    \label{eq: total CC for all methods}
\end{equation}
Note that the number of additional multiplications grows linearly with the size of the ESN, the total number of multiplications grows quadratically and $N>>1$, so the quadratic term dominates in the complexity. We used the classical training procedure as described in~\cite{Jaeger2004}. This procedure together with the regularization aspects are given in Supplementary materials.

\begin{figure}[ht!]
\includegraphics[width=.98\linewidth]{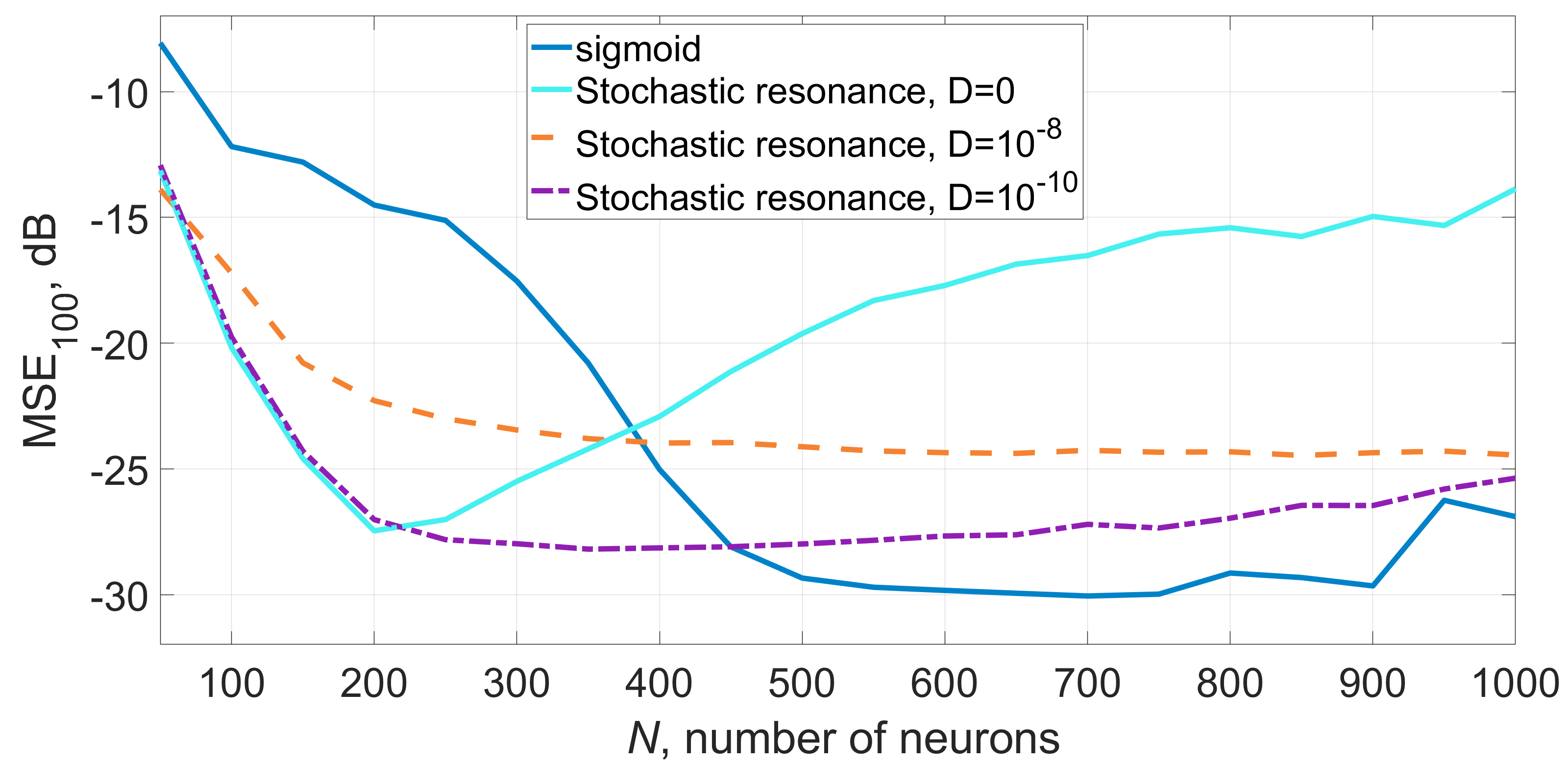}
\caption{ Mean squared error (MSE) for different nonlinear activation functions depending on the number of neurons.}
\label{fig: MSE sigmoid vs noisy SR}
\end{figure}

The prediction accuracy is measured by the mean squared error between freely running ESN and corresponding 100 samples of the MG series.  
Parameters of the ESN under test are as below:
\begin{itemize}
\item $N=50\cdots1000$
\item Connectivity of $\mathbf{W}$ is 0.01
\item Hyperparameter of the SR function $\alpha=0.01$
\item SR noise amplitude $D=0, 10^{-10}, 10^{-8}, 10^{-6}$
\end{itemize}
The result of this test is shown in Fig.~\ref{fig: MSE sigmoid vs noisy SR} where each point represents a mean value averaged over 1000 samples. 

The "classical" ESN with sigmoid activation function is trained under similar conditions. The linear regression problem for determining the readout weights was performed using singular value matrix decomposition. Other aspects of the training procedure and regularization are given in Supplementary material.

\begin{figure}[ht!]
\includegraphics[width=.98\linewidth]{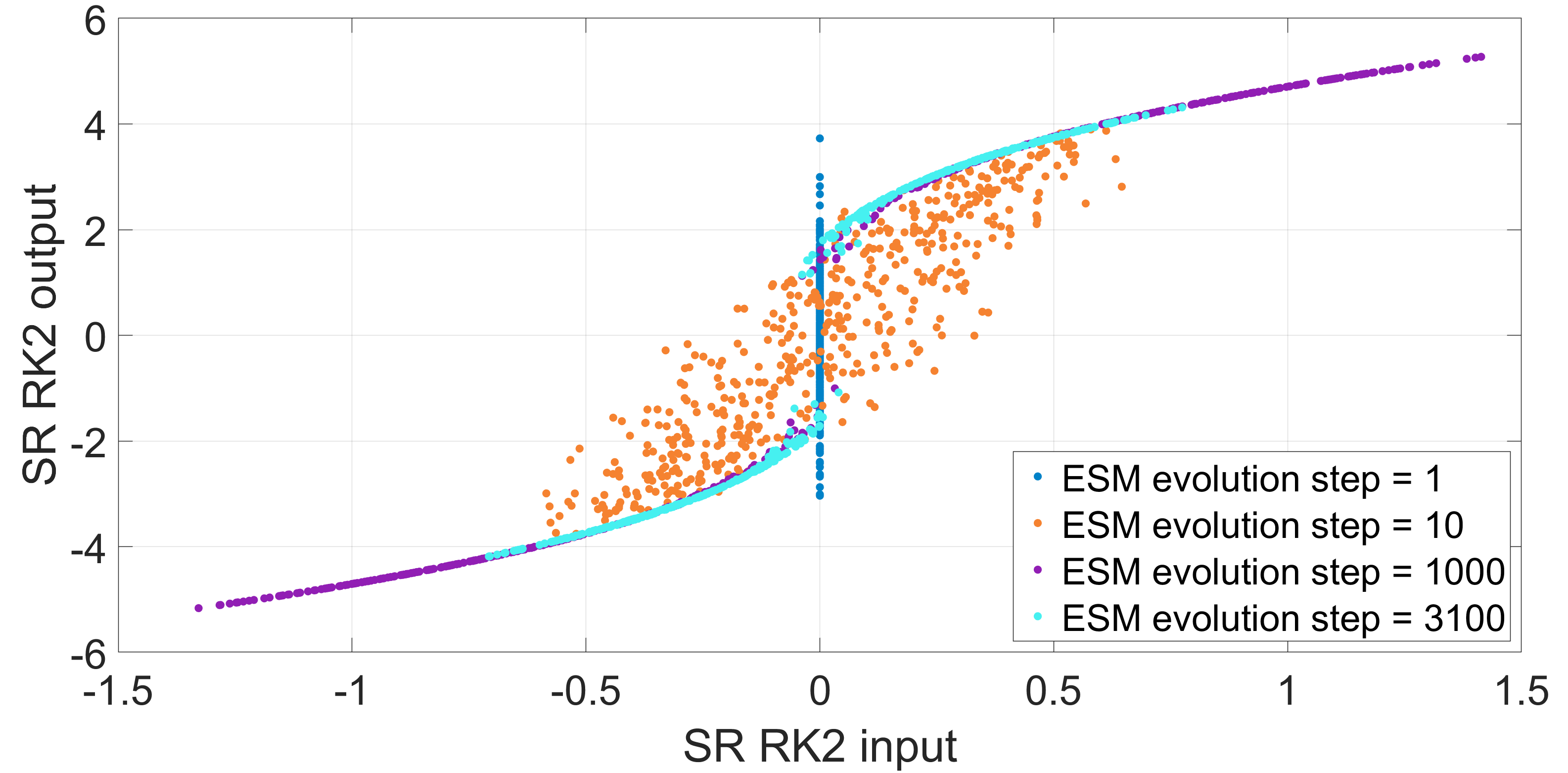}
\caption{Stochastic resonance transfer function}
\label{fig: SR RK2 transfer function}
\end{figure}

\begin{figure}[ht!]
\includegraphics[width=.98\linewidth]{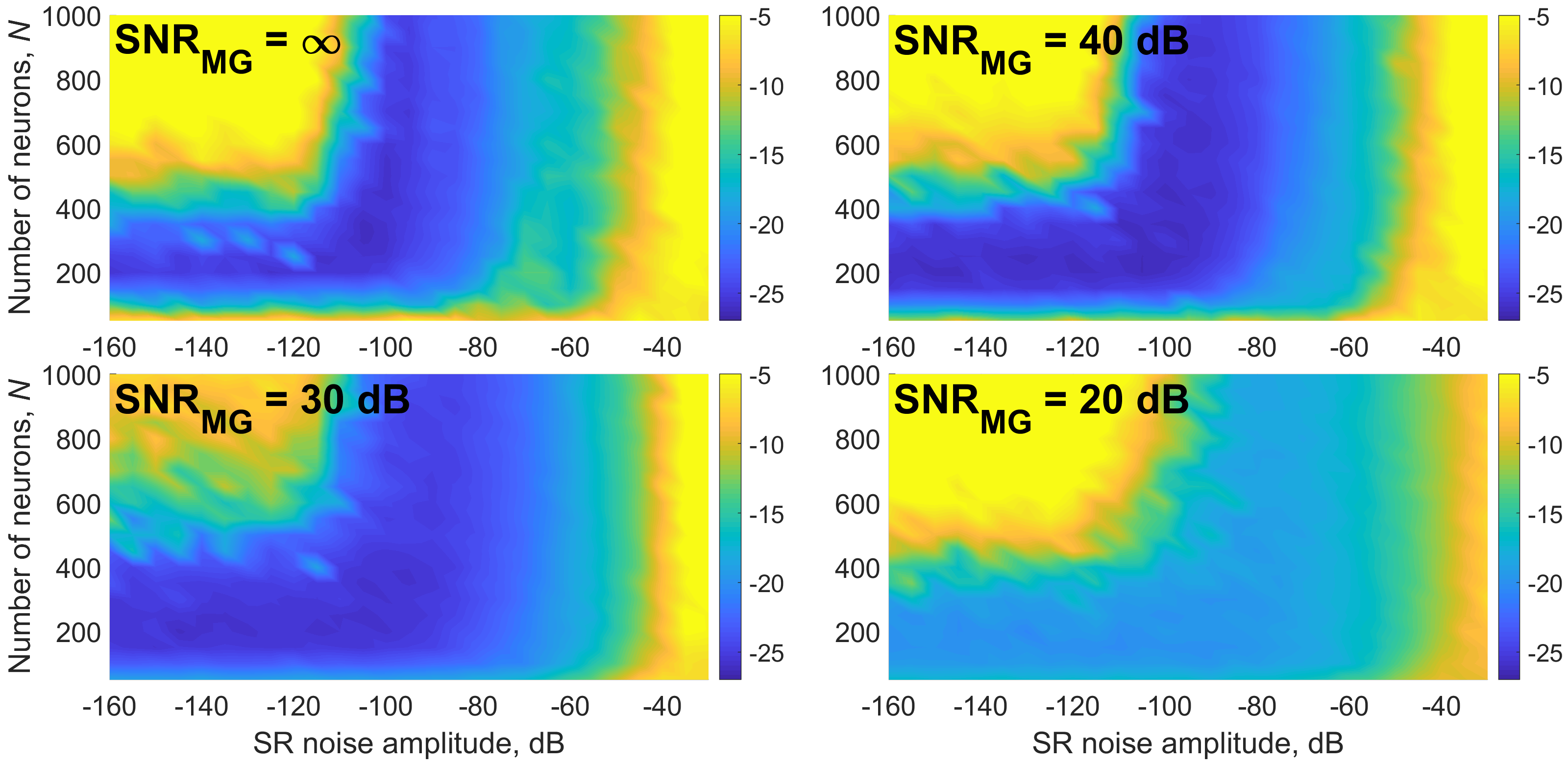}
\caption{Prediction accuracy (colour-coded) depending on SR noise level, number of neurons and noise in the teaching sequence.}
\label{fig: optimal SR noise level}
\end{figure}

The transfer functions of an SR neuron depends on the number of the ESN evolution step. The initial values (step 1) are normally distributed with mean 0 and variance 1.  The transfer function for steps 1, 10, 1000 and 3100 are shown in Fig.~\ref{fig: SR RK2 transfer function}. Each neuron automatically converges into its own transfer function during the learning procedure, providing a possibility of the self-adjusting activation functions in different layers of ESN using the same design of the node.
One can see that for lower number of neurons the performance of ESN-SR is better than the classical approach with a sigmoid activation function. In particular, the SR method reaches its maximum accuracy at $N=200$ neurons with an averaged error of $10^{-2.76}$. The number of multiplications per 1 step is $Q_{\text{total}}^{SR}(N=200)=40800$. For the classical sigmoid an error of $10^{-1.45}$ or 0.036 is achieved for a similar number of nodes $Q_{\text{total}}^{\text{classical}}(N=200)=40600$. So we obtain 20 times more accurate results by using the ESN of same computational complexity.
To achieve the same accuracy using the sigmoid activation function one needs to take $N=450$ neurons leading to $Q_{\text{total}}^{\text{classical}}(N=450)=203850$ multiplications per 1 step. So, to achieve the same accuracy using the classical sigmoid function one needs to perform 5 times more multiplications and 2.5 times more nodes.
\comment{
It should be mentioned that the maximum accuracy achieved by using the sigmoid function is slightly higher, but at the expense of much more computational cost: the maximum accuracy of $10^{-3}$ is achieved at  $N=700$ neurons that require $Q_{\text{total}}^{\text{sigmoid}}(N=700)=492100$ multiplications per 1 step. So the best result is only $1.75$ times better compared to the best SR result while it is achieved by using a 12 times more multiplications.\\
}
We investigated how the noise amplitude $D$ affects the accuracy and stability of the ESN, see Figs~\ref{fig: MSE sigmoid vs noisy SR} and \ref{fig: optimal SR noise level}. One can see that the low level of added noise improves the maximum accuracy slightly and also increases stability and accuracy at a higher number of neurons by preventing overfitting. But even slightly higher noise amplitude reduces the accuracy. 
We also investigated the capability of predicting the continuation of the MG series when learning on noisy sequence. The training sequence was corrupted by white Gaussian noise and fed into the ESN with the same learning procedure as before. \\
The dependence of the MSE (colour-coded) on number of neurons, internal noise in the nonlinear activation function and SNR of the teaching sequence is shown in Fig.~\ref{fig: optimal SR noise level} showing the optimal internal SR noise level close to $10^{-10}$.\\
We used the SR function with $D=10^{-10}$ and compared it with classical sigmoid function. Noise amplitudes in the training sequence $\sigma$ corresponding to SNR of $20$, $30$ and $40$ dB were chosen. Figure~\ref{fig: MSE RK2 vs sigmoid, noisy MG} shows how the MSE of the first 100 predicted values depends on the number of neurons for different nonlinear activation functions and various noise levels in the training sequence.
\begin{figure}[ht!]
\includegraphics[width=.98\linewidth]{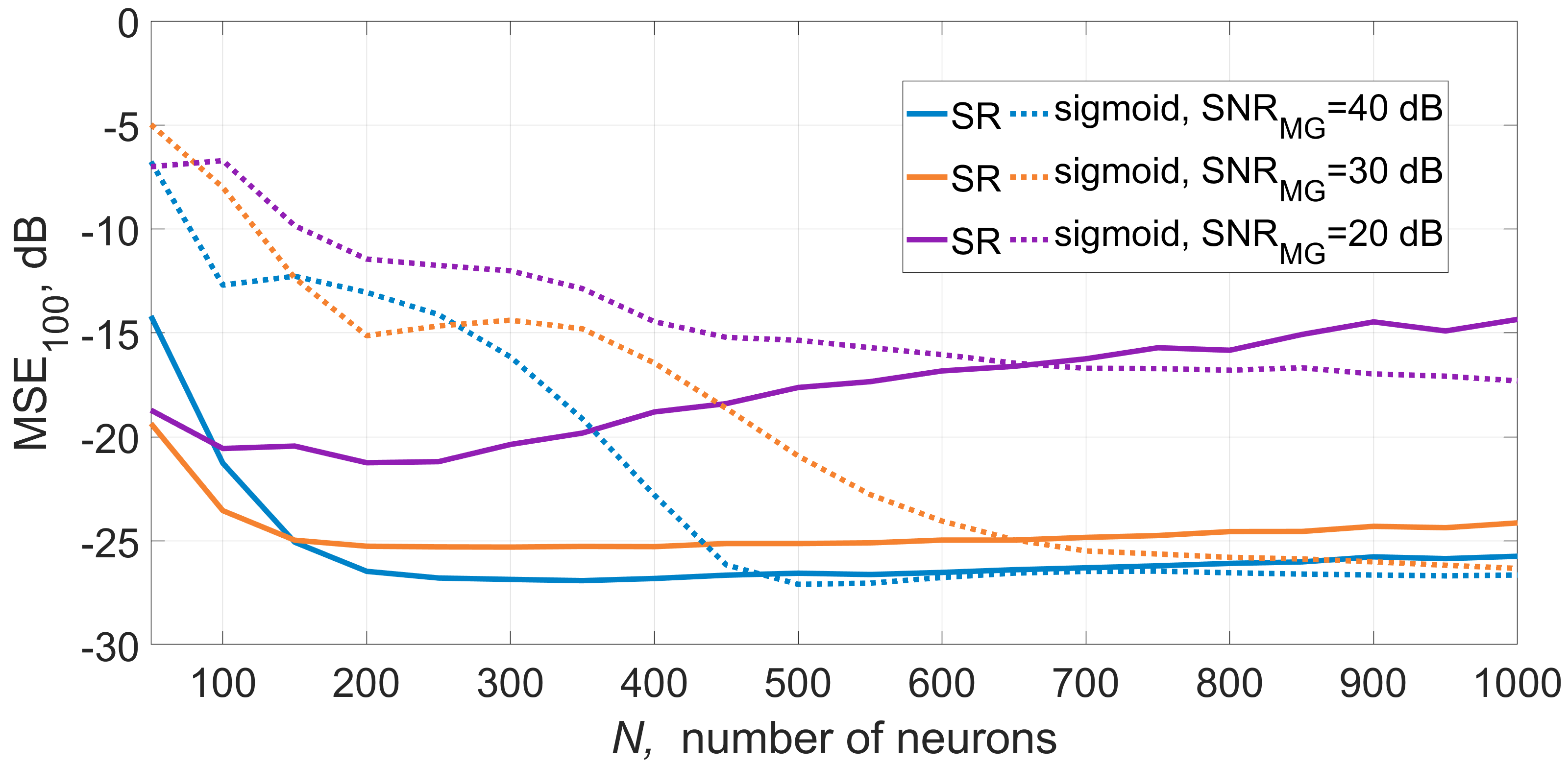}
\caption{Accuracy of the RC when learning on noisy MG sequence.}
\label{fig: MSE RK2 vs sigmoid, noisy MG}
\end{figure}
The proposed method shows superior performance compared to classical approach in case of lower number of neurons and the same performance for higher number of neurons. In particular, is case of SNR~$=20$ dB the prediction accuracy is as good as $0.01$ when the number of neurons is as low as 
$100$ in case of SR. The use of sigmoid function provides 24 times less accurate results at this number of neurons. And this accuracy is never achieved with the classical sigmoid function even at higher number of neurons at this level of noise in the training sequence.
Figure~\ref{fig: best acc SRRK2 vs sigmoid} shows how the best prediction accuracy across various number of neurons depends on the SNR in the training sequence where shaded regions depict one standard deviation intervals calculated on 1000 runs.
\begin{figure}[ht!]
\includegraphics[width=.98\linewidth]{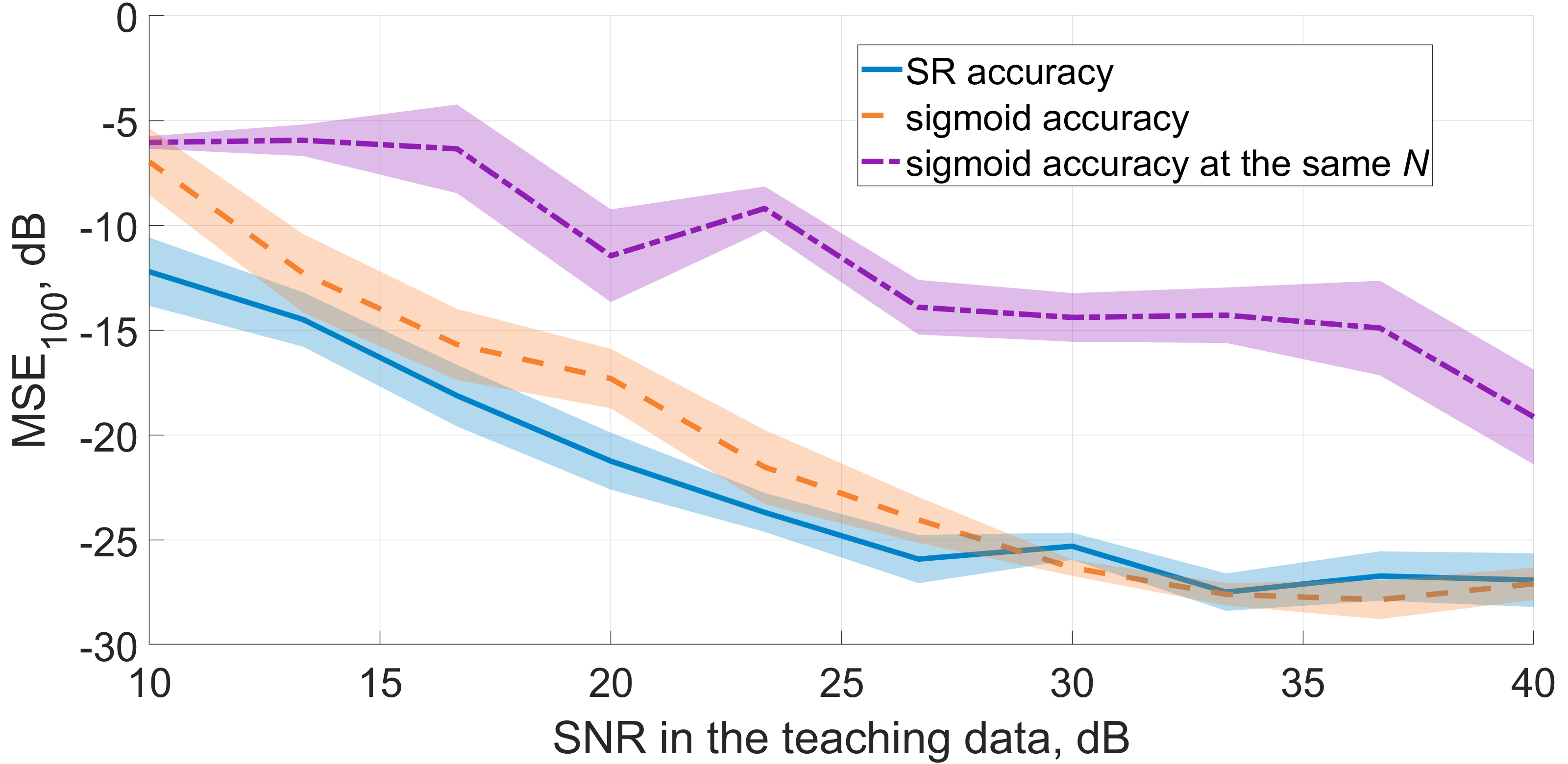}
\caption{Best accuracy across ESNs of various complexity. Shaded areas depict one standard deviation intervals. }
\label{fig: best acc SRRK2 vs sigmoid}
\end{figure}
As can be seen in Fig.~\ref{fig: best acc SRRK2 vs sigmoid} while there is no statistically significant difference between the performance of the ESN-SR and sigmoid system for SNR $>25$~dB, the latter outperforms the former in other regions especially when the number of nodes is the same (violet curve). We believe this result is particularly important for training on experimental data as it is always corrupted by noise.\\
\comment{
We have shown that replacing the standard non-linear function with SR significantly increased the accuracy of the ESN with fewer neurons when trained on a Mackey-Glass series. The proposed approach requires 2 times fewer neurons than classical one to achieve an error of 0.0015: the computational complexity in this case has decreased by a factor of 4. 
Moreover, since noise is ubiquitous in any real implementation of an ESN, a more realistic investigation requires considering the impact of noise introduced to the system through the nonlinear activation node. Our simulations show that ESN-SR outperforms the one with conventional nodes by a margin that is dependent on the noise power. With a low noise level, the accuracy of the ESN is slightly improved and overfitting is significantly reduced for a larger number of neurons. 
When training on noisy data, the proposed approach is significantly superior to the classical one both in accuracy and computational complexity. When SNR~$=40$ dBm and with the same computational complexity, ESN-SR achieves a 25x accuracy improvement compared to the ESN-sigmoid. To deliver the same accuracy, the classical approach requires more than 4 times more calculations.\\
This difference becomes more prominent for noisier data: for SNR $=20$ dB with the same computational complexity, the best achieved accuracy of the proposed method is $9.5$ times better than the one with sigmoid nodes. At the same time, the maximum accuracy of the classical approach across all simulated number of neurons, is more than $2.5$ times lower and requires 25 times more calculations compared to the ESN-SR.\\
}
In this work we proposed using SR as a nonlinear activation function in neural networks and investigated how SR nodes can improve the performance of an ESN. Using standard Mackey-Glass equation benchmark and ESN as an illustration of the general concept, we demonstrated the superiority of an ESN-SR computer to ESN-sigmoid in the case of noisy input data and implementation noise. We showed that our proposed ESN can provide up to 9.5 times better accuracy in predicting the MG time series compared to the conventional ESN with the same number of nodes. This indicates the capability of SR nodes to capture the underlying relations between samples of the input and manifesting memory properties in various ANNs. 
We believe that the proposed idea of using model (or, indeed, physical systems) governed by stochastic ordinary differential equations can be applied in a range of ANNs and can be generalized to different tasks. In particular, the proposed concept is compatible with high-bandwidth optical analogue ANNs and reservoirs offering potential solutions for high-speed parallel signal processing and reduction in the power consumption in physical implementations.

This work was supported by the EU ITN project POST-DIGITAL and the EPSRC project TRANSNET.\\
\bibliography{Bibliography.bib}
\end{document}


\newcommand{\comment}[1]{}

\preprint{APS/123-QED}

\title{Supplementary materials: \\ Stochastic resonance neurons in artificial neural networks}

\author{E. Manuylovich, D. Arg\"uello Ron, M. Kamalian-Kopae, S. K. Turitsyn }

\address{Aston Institute of Photonic Technologies, Aston University, Birmingham B4 7ET, UK}
\date{\today}


\begin{abstract}

\end{abstract}


\maketitle


%
%


\section{Mackey-Glass series}

To evaluate the performance of the SR as the nonlinear function we used a classical problem of predicting the Mackey-Glass (MG) series described by the following equation:
\begin{equation} 
\begin{aligned}
& \dot{q}(t)=\frac{aq(t-\tau)}{1+q^{10}(t-\tau)}-bq(t) \\
\end{aligned}
    \label{eq: MG equation}
\end{equation}
with parameters $a=0.2$, $b=0.1$ and $\tau=17$. We integrated this equation using Matlab delay differential equations solver dde23 over a time interval $t\in(0,3000)$ with an initial value $q(0)=0.5$, then we interpolated the solution into a grid $t=[50:\Delta t:3000]$ with $N_t=3500$ points. Corresponding time interval $\Delta t=2950/3500\approx0.843$. The initial interval $t\in(0,50)$ was discarded to eliminate the initial transitions of the system from the training sequence. 
The MG series used to train and test ESNs is shown in Fig.~\ref{fig: MG series} 
\begin{figure}[h!]
\centering\includegraphics[width=1\linewidth]{SMfig01. MG series and ESN prediction.png}
\caption{The MG series is used to train the ESN and estimate its accuracy.}
\label{fig: MG series}
\end{figure}

\comment{
\section{Computational complexity}

To investigate the computational complexity of the proposed approach we calculate the number of multiplications required for a single step of ESN evolution. This number includes both linear part of the ESN evolution and the multiplications needed for calculating the nonlinear response of the neurons. 

The evolution of the ESN is described by the following equations:
\begin{equation} 
\begin{aligned}
& x_{n+1} = f(\mathbf{W}_{in}u_{n+1} +  \mathbf{W}x_{n} + \mathbf{W}_{back}y_{n+1}) &\\
\end{aligned}
\label{eq: ESN evolution step}
\end{equation}

\begin{equation} \label{eq: ESN readout layer equation}
y_{n+1} = \mathbf{W}_{out} x_{n+1}  
\end{equation}

According to equations ~\ref{eq: ESN evolution step} and ~\ref{eq: ESN readout layer equation}, the linear part of ESN evolution includes 4 matrix multiplications of size $N\!\!\times\!\!1$, $N\!\!\times\!\! N$, $N\!\!\times\!\!1$, and $1\!\!\times\!\!N$ giving a total of
\begin{equation} 
\begin{aligned}
& Q_{\text{linear}} = N^{2}+3N
\end{aligned}
    \label{eq: CC of linear part}
\end{equation}
multiplications per 1 step of linear part of the ESN evolution. This is also the total number of multiplications for the classical approach for the nonlinear activation function such as sigmoid or tanh is obtained using lookup table with no computational complexity.

On the other hand, for the SR nonlinear function, the values of $\alpha(\xi-\xi^{3})+D\sigma(t)$ or $\left[\alpha(\xi_n-\xi_n^{3})+D\sigma(t)\right]\cdot\Delta t$ can also be calculated using look-up tables and bear no computational burden. However, $s_n \cdot \Delta t$ needs to be calculated and add up $\xi_n$, $\left[\alpha(\xi_n-\xi_n^{3})+D\sigma(t)\right]\cdot\Delta t$, and  $s_n \cdot \Delta t$. So an additional
\begin{equation}     \label{eq: SR RK2 nonlinear CC}
Q_{\text{nonlinear}}^{SR}=N 
\end{equation}
multiplications are needed to calculate the SR nonlinear response function.

To sum up, the total numbers of multiplications for 1 step of calculating evolution of the ESN are:
\begin{equation} 
\begin{aligned}
& Q_{\text{total}}^{\text{classical}}=N^2+3N \\
& Q_{\text{total}}^{SR}=N^2+4N \\
\end{aligned}
    \label{eq: total CC for all methods}
\end{equation}
Note that the number of additional multiplications grows linearly with the size of the ESN, the total number of multiplications grows quadratically and $N>>1$, so the quadratic dominates in the complexity.
}

\section{Training procedure}

We used the classical training procedure as described in \cite{Jaeger2004}. This procedure is as follows:
\begin{enumerate}
\item Feed the first 3000 elements of the Mackey-Glass series into the output of the RC
The feeding is the following procedure:
\begin{equation} 
\begin{aligned}
& y_{n+1}=q_{n+1} \\
& x_{n+1}=f(\mathbf{W}x_n+\mathbf{W}_{back}y_n)
\end{aligned}
\end{equation}
During this the RC must start “echoing” the fed signal: after some transitional period, the values of internal weights $x$ are starting to show some variations of the fed signal.
\item Use the last 2000 internal states   to train the output weights so at each step the product of the output weights matrix and the internal state vector gives the corresponding output value:
\begin{equation} 
\begin{aligned}
\mathbf{W}_{out}=\arg\min_{\mathbf{W}_{out}} \sum|\mathbf{X}\mathbf{W}_{out}-y|^2
\label{eq: LMSE training problem}
\end{aligned}
\end{equation}
Here $\mathbf{X}$ is a matrix composed of the last 2000 internal states $x$. $\mathbf{W}_{out}$ can be determined by finding minimum norm least-squares solution to the linear equation.
\end{enumerate}
Training is complete at this stage.

Then the training sequence is disconnected from the RC and the RC is running freely. The next several output values produced by the freely running ESN are compared to the true values produced by the Mackey-Glass equation.
The prediction accuracy is determined by the mean squared error between the first 100 samples of the MG series and the freely running ESN. 

\section{Regularization}
The training procedure for any of the considered ESNs requires solving the least mean square problem \ref{eq: LMSE training problem}. The analytical solution to this problem is well known:
\begin{equation} 
\begin{aligned}
\mathbf{W}_{out} = \left(\mathbf{X}^T\mathbf{X}\right)^{-1}\mathbf{X}^T y
\label{eq: analytical solution to non-regularized LMSE}
\end{aligned}
\end{equation}

One could, in principle, use this analytical solution to solve problem \ref{eq: LMSE training problem} and \textit{calculate} the weight vector $W_{out}$. However, in a common practical case of multicollinearity of columns of matrix $\mathbf{X}$, the matrix $\mathbf{X}^{T}\mathbf{X}$ becomes ill-conditioned. In this case, the Cholesky decomposition, that is the fastest and the standard way for calculating inverse matrices, becomes unstable and cannot provide accurate results. As a consequence, the resulting weight vector of Eq.~(\ref{eq: analytical solution to non-regularized LMSE}) does not provide the LMSE solution and the prediction accuracy dramatically drops.

There are several strategies to overcome this problem one of which is the well-known approach described in \cite{lukovsevivcius2009reservoir}, called ridge regression or ridge regularization.

In this case, a regularization term is included in the minimization problem:
\begin{equation} 
\begin{aligned}
\mathbf{W}_{out}=\arg\min_{\mathbf{W}_{out}} \left(\sum|\mathbf{X}\mathbf{W}_{out}-y|^2 + \lambda||{W}_{out}||^2\right)
\end{aligned}
\end{equation}
where the norm of the vector ${W}_{out}$ is taken into account alongside the mean square error.

This problem is equivalent to the following LMSE problem:
\begin{equation} 
\begin{aligned}
\left(\mathbf{X}^T\mathbf{X} + \lambda\mathbf{I}\right)\mathbf{W}_{out} - \mathbf{X}^T y = 0
\label{eq: ridge regularized LMSE problem}
\end{aligned}
\end{equation}
the exact analytical solution of which is  
\begin{equation} 
\begin{aligned}
\mathbf{W}_{out} = \left(\mathbf{X}^T\mathbf{X} + \lambda\mathbf{I}\right)^{-1}\mathbf{X}^T y
\label{eq: analytical solution to ridge regularized}
\end{aligned}
\end{equation}
Here $\mathbf{I}$ is a an identity matrix of the same size as $\mathbf{X}^T\mathbf{X}$.
The main advantage of this method is that the matrix to be inverted, $\mathbf{X}^{T}\mathbf{X}+\lambda \mathbf{I}$ can be made well-conditioned and therefore can be inverted using standard and fast Cholesky decomposition.
For this regularization method one should carefully tune the value of the parameter $\lambda$ as for very small values it does not provide any regularization and matrix to be inverted is still ill-conditioned. And for high values of $\lambda$ this problem is too far from the original one, so the obtained solution does not catch the true relations between the internal state of the ESN and the observed teaching sequence.

Another technique for solving \ref{eq: LMSE training problem} also mentioned in \cite{lukovsevivcius2009reservoir} is singular value decomposition (SVD) where the authors emphasis on its stability and robustness. It can provide the minimum norm least-squares solution to the problem \ref{eq: LMSE training problem} even in case of multicollinearity and for ill-conditioned matrix $\mathbf{X}$, however, it is more computationally expensive.

Another standard method for solving \ref{eq: LMSE training problem} is the QR decomposition which is as computationally stable as SVD, but less computationally expensive.

Here we compare these three approaches as applied to the linear regression problem and therefore their ability to be trained and to capture the underlying relations between the target and the input data.

We apply all 3 methods described above to solve equation \ref{eq: ridge regularized LMSE problem} to find out how regularization affects the performance. In order to do that, we calculate the prediction accuracy depending on the ridge regularization parameter for the described approaches.
First we construct a matrix $\mathbf{X}$ of size $m\times n$ and of rank $k$. We chose "tall" matrix $\mathbf{X}$ with $m>n$ so it corresponds to the ESN training procedure where we use 2000 observations to train 1000 or less neurons. We deliberately chose rank deficit matrix $X$ with $k<n$ to provide multicollinearity in columns of the matrix $\mathbf{X}$. We construct the random matrix $\mathbf{X}=\mathbf{U}\mathbf{S}\mathbf{V}'$ using the following formula:
where $\mathbf{U}$ is an $m\times m$ orthogonal matrix with random normally distributed elements, $U_{ij}\sim\mathcal{N}(0,1)$; $\mathbf{V}$ is an $n$ by $n$ matrix of the same type, and $\mathbf{S}$ is an $m$ by $n$ matrix with only $k$ ($k<n<m$) non-zero normally distributed random elements on the main diagonal $S_{ii}\sim\mathcal{N}(0,1)$, $i\le k$. This ensures that the matrix $\mathbf{X}$ has the desired properties.

Then we split this matrix horizontally in two parts: $\mathbf{X}_{train}$ and $\mathbf{X}_{test}$ with the number of rows in the training matrix $t$, larger than $k$.

Then we construct a random normally distributed vector $W_j\sim\mathcal{N}(0,1)$ of true weights which we then use to calculate the training and testing answers:

\begin{equation} 
\begin{aligned}
y_{train} = \mathbf{X}_{train}W
\label{eq: train answers}
\end{aligned}
\end{equation}
\begin{equation} 
\begin{aligned}
y_{test} = \mathbf{X}_{test}W
\label{eq: test answers}
\end{aligned}
\end{equation}

Now we finally solve the linear regression problem on training data using 3 methods described above and test these solutions by calculating MSE on testing data.
The error is calculated as
\begin{equation} 
\begin{aligned}
error_{train/test} = \frac{||\mathbf{X}_{train/test}W-y_{train/test}||}{||y_{train/test}||}
\label{eq: training testing errors}
\end{aligned}
\end{equation}

Figure \ref{fig: analytical ridge vs QR vs SVD} shows how training and testing error depend on the ridge regularization parameter for various methods.
\begin{figure}[h!]
\centering\includegraphics[width=1\linewidth]{SMfig02. Analytical ridge vs QR vs SVD.png}
\caption{Comparison of training and testing errors depending on regularization parameter for various methods.}
\label{fig: analytical ridge vs QR vs SVD}
\end{figure}

One can see that even though the classical ridge regularization technique works for some values of the regularization parameter $\lambda$, it is much worse in terms of prediction accuracy even when the regularization parameter is carefully optimized. At the same time, the methods based on QR or SVD decomposition show great prediction accuracy at very low values of the regularization parameter that means they work fine without any regularization at all. 

To confirm that we applied the latter two methods directly to problem \ref{eq: LMSE training problem} and found out that even for ill-conditioned matrix $\mathbf{X}$ with multicollinear columns both training and testing errors for these methods are of order of $10^{-15}$ with no regularization at all.

Therefore, in our work we chose the SVD-decomposition-based method with no regularization for solving LMSE \ref{eq: LMSE training problem} instead of analytical solution with ridge regularization even though it has higher computational cost. Since training must be performed only once, this higher computational complexity is acceptable. For larger matrices one can use QR-decomposition-based method as it provides the same accuracy at lower computational costs compared to the SVD. In our case the processing time difference between QR and SVD was negligible so we used SVD as it provides minimum norm solution.

\comment{
regularization to avoid overfitting and ensure the optima performance. This is important because we need to compare the newly developed SR ESN with properly trained classical one.\\
The training procedure requires solving the linear LMSE problem 
\begin{equation} 
\begin{aligned}
\mathbf{X}\mathbf{W}_{out} - y = 0 
\end{aligned}
\end{equation}
which standard solution is well known:
\begin{equation} 
\begin{aligned}
\mathbf{W}_{out} = \left(\mathbf{X}^T\mathbf{X}\right)^{-1}\mathbf{X}^T y
\end{aligned}
\end{equation}
However, if the matrix $\mathbf{X}^T\mathbf{X}$ is ill-conditioned due to multicollinearity of the eigenvectors of matrix $\mathbf{X}$, then the solution of this equation becomes unstable: the vector $\mathbf{W}_{out}$ is not then defined by the true underlying relations between the matrix $\mathbf{X}$ and observed vector $y$. This leads to a significant drop in the prediction accuracy. 
\\
In order to avoid the ill-conditioned matrix inversion, multiple regularization techniques can be applied. One of the techniques (see \cite{Jaeger2004}) consists of adding just a bit of  noise to the output of each neuron so that the initially highly correlated columns in $\mathbf{X}$ become much less correlated. The typical value of $10^{-10}$ can be used \cite{Jaeger2004}. Then the minimum norm least-squares solution to linear equation can be found to reconstruct the main components of the reservoir that correspond to the observed teaching sequence. This is the main method that we used for regularization of the learning process for the "standard" sigmoid-based ESN.
Another way or regularizing the learning process is ridge regression. In this case instead of minimizing $\sum|\mathbf{X}\mathbf{W}_{out}-y|^2$ one can solve a similar problem 
\begin{equation} 
\begin{aligned}
\mathbf{W}_{out}=\arg\min_{\mathbf{W}_{out}} \left(\sum|\mathbf{X}\mathbf{W}_{out}-y|^2 + \lambda||{W}_{out}||^2\right)
\end{aligned}
\end{equation}
In this problem statement we not only aim to minimize the mean square error, but also the norm of the vector ${W}_{out}$. \\
The exact solution to this problem is  
\begin{equation} 
\begin{aligned}
\mathbf{W}_{out} = \left(\mathbf{X}^T\mathbf{X} + \lambda\mathbf{I}\right)^{-1}\mathbf{X}^T y
\end{aligned}
\end{equation}
Here $\mathbf{I}$ is a an identity matrix of the same size as $\mathbf{X}^T\mathbf{X}$.
For this regularization method one should carefully tune the value of the parameter $\lambda$ as for very small values it does not provide any regularization and matrix to be inverted is still ill-conditioned. And for high values of $\lambda$ this problem is too far from the original one, so the obtained solution does not catch the true relations between the internal state of the ESN and the observed teaching sequence. We tried both regularization techniques and found out that the first one works better, that is why we used it.

Without any regularization the test error rises quickly due to overfitting with the increase of the number of neurons. So another sign of proper regularization is that the test error does not rise with the increased number of neurons in ESN. We ensured that this condition is sutisfied for the "classical" sigmoid-based ESN that we used for our tests.
\\
\section{A few more words on regularization}
Herbert refers to this work of theirs:\cite{lukovsevivcius2009reservoir} when he says about need for accurate ridge regularization.

There is the paragraph 8.1.1. Linear regression about regularization. In particular, the "classical" analytical approach to solving the problem of linear regression in the sense of least squares is considered there, when the solution to the linear regression problem
\begin{equation} 
\mathbf{X} W_{out}=y
\label{eq: linear regression problem}
\end{equation}

in the sense of least squares
\begin{equation} 
W_{out}=\arg\min_{W_{out}} \left(\sum|\mathbf{X}W_{out}-y|^2\right)
\label{eq: LMSE definition}
\end{equation}

is written analytically:
\begin{equation} 
W_{out}=\left( \mathbf{X}^{T}\mathbf{X} \right)^{-1}\mathbf{X}^{T}y
\label{eq: analytical solution to regression problem}
\end{equation}

This solution, although analytically correct, is absolutely useless in the case of an ill-conditioned matrix $\mathbf{X}$, since the inverse matrix on the right-hand side cannot be calculated, and the solution to problem \ref{eq: LMSE definition} turns out to be unstable.
In the article above, Herbert gives 2 possible ways: solving the problem \ref{eq: linear regression problem} using the SVD decomposition of the $\mathbf{X}$ matrix and calculating the pseudo-inverse matrix, or solving the problem using ridge regression, when problem \ref{eq: LMSE definition} is replaced by this one:

\begin{equation} 
W_{out}=\arg\min_{W_{out}} \left(\sum|\mathbf{X}W_{out}-y|^2 + \lambda||W_{out}||\right)
\label{eq: ridge LMSE problem}
\end{equation}

And the corresponding solution is:
\begin{equation} 
W_{out}=\left( \mathbf{X}^{T}\mathbf{X} + \lambda\mathbf{I}\right)^{-1}\mathbf{X}^{T}y
\label{eq: ridge analytical solution}
\end{equation}

Herbert argues that computing a pseudoinverse matrix is a good, robust method, but computationally expensive. So, he says, let's focus on ridge regression. And then comes the description of regularization by the ridge regression method. He does not mention the QR decomposition of the matrix $\mathbf{X}$ for solving the linear regression problem, although it is just as stable as SVD, but less computationally expensive.

For the test, I solved problem \ref{eq: linear regression problem} with some random matrices $\mathbf{X}$ and $y$, however, I deliberately made the problem similar to our case with SR: I made the $\mathbf{X}$ matrix "tall", that is, the number of observations is greater than the number of features (over-determined problem, solution possible only in the sense of least squares), but at the same time with a rank deficit (the number of linearly independent columns is less than the number of columns)

I compared the solutions to problem \ref{eq: linear regression problem} by the analytical method with a ridge regularizer, QR decomposition with a ridge regularizer, and SVD decomposition with a ridge regularizer. See figure \ref{fig: analytical ridge vs QR vs SVD} for the comparison of these methods.

It turns out that for the analytical solution there is some optimal value of the regularization parameter at which the best approximation is observed. This is fully consistent with what Herbert said.

At the same time, and this is the key point, for QR decomposition and SVD decomposition, even with a near-zero value of the parameter $\lambda$, a good solution and a good approximation are still obtained. The logarithm of the root-mean-square error is at the level of $10^{-15}$ even in the absence of regularization (at $\lambda = 10^{-20}$), and the norm of the weight vector is about unity, which means that there is no overfitting.

In general, everything works fine even without ridge regularization, if we use not an analytical solution (which gives an unstable solution due to the singularity of the matrix), but QR or SVD decomposition methods.}

\bibliography{Bibliography.bib}